\documentclass[a4paper]{article}
\usepackage{inputenc}
\usepackage{graphicx}
\usepackage{siunitx}
\usepackage{onecolceurws}
\usepackage{hyperref}
\usepackage{verbatim}
\usepackage{subfig}
\usepackage{algorithm}
\usepackage[noend]{algpseudocode}
\graphicspath{ {img/} }

\usepackage{apalike}
\hypersetup{
    colorlinks=false,
    pdfborder={0 0 0},
}
\title{Adaptive Probabilistic Tack Manoeuvre Decision for Sailing Vessels}

\author{
S\'ebastien Lemaire \\ sebastien.lemaire@soton.ac.uk
\and
Yu Cao \\ Yu.Cao@soton.ac.uk
\and
Thomas Kluyver \\ thomas@kluyver.me.uk \vspace{0.6cm}
\and
Daniel Hausner \\ dh4n16@soton.ac.uk
\and
Camil Vasilovici \\ crv1g16@soton.ac.uk 
\and
Zhong-yuen Lee \\ leezhongyuen@gmail.com\vspace{0.6cm}
\and
Umberto Jos\'e Varbaro \\ ujv1u16@soton.ac.uk
\and
Sophia M. Schillai \\ sms4g13@soton.ac.uk
}

\institution{University of Southampton \\ 
Boldrewood Innovation Campus\\ Southampton, S016 7QF}

\begin{document}
\maketitle

\begin{abstract}
    To move upwind, sailing vessels have to cross the wind by tacking. During
    this manoeuvre distance made good may be lost and especially smaller vessels
    may struggle to complete a tack in averse wind and wave conditions.
    A decision for the best tack manoeuvre needs to be made based on weather and
    available tack implementations. 
    
    This paper develops an adaptive probabilistic tack manoeuvre decision
    method. The order of attempting different tacking strategies is based on
    previous success within a timeout, combined with an exploration component.
    This method is successfully demonstrated on
    the 1m long sailing vessel Black Python. Four strategies for crossing the wind
    were evaluated through adaptive probabilistic choices, and the best 
    was identified without detailed sensory knowledge of the actual weather conditions.

    Based on the positive results, further improvements for a better selection
    process are suggested and the potential of using the collected data to
    recognise the impact of weather conditions on tacking efforts is recognised.
\end{abstract}
\vskip 32pt

\newpage
\section{Introduction}
  Current work on manoeuvre planning for sailing robots focuses on long term
  piloting from planning the actions until the next waypoint to routing a vessel
  over longer distances \cite{tynan2018attractor,langbein2011rule}.
  When the routing towards a waypoint, all sailing vessels rely on
  manoeuvres to cross the wind: tacking, where the vessel turns with the bow
  facing towards the wind, or jibing, where the vessel turns with the bow facing
  away from the wind. 
  Typically, sailing robots are controlled by using a rudder to control the
  heading of the vessel and then adjusting the sails based on the relative wind
  direction \cite{gomes2017improving} even if recent sail designs may include heading control in the sail
  \cite{augenstein2017using}.
  Vessel speed and boat drag are often recognised as a factor for successfully completing a
  tack \cite{jouffroy2009control,jouffroy2009steering}.
  \cite{cruz2014navigation} recognises that a sailing robot can get
  stuck facing into the 
  wind, in sailing terms 'in irons', and suggests to recover 
  from such a situation by increasing speed; thus gaining rudder control through letting the sails loose 
  so the wind can push the boat backwards. 
  \cite{tranzatto2015debut} studies how to perform fast and smooth tack manoeuvres using control system
  theory and compares 3 different rudder controls for tacking, however tack fails
  are not mentioned.
  Modeling a tacking manoeuvre could also be done to analyse the problem of
  tack failure. Several
  studies developed tacking simulators based partially on experimental
  measurements \cite{masuyama2011tacking,roncin2004dynamic,spenkuch2010real}, however they are applied on
  large sailing boats where tack failure is not an issue, and is hence
  not discussed.
  When sailing the 1m long Southampton Sailing Robot, the Black Python, we found that the success
  and speed of a tack manoeuvre for such a small vessel not only depends 
  on having sufficient speed and suitable rudder action to
  pass through the wind, but also on passing through the wave fronts pushed
  towards it by the wind. 
  Where a human sailor would make choices about the sail
  and rudder settings based on speed and wave observations and
  experience, making small adjustment as the manoeuvre proceeds,
  implementing this process for a robotic sailor is challenging. Not
  only is it difficult to translate experience into software, but also
  the amount of sensor data that is required increases significantly
  compared to a dead-reckoning tack manoeuvre. 
  As an alternative to fully measuring and considering all factors involved, the
  introduced system adapts to the wind and waves conditions by testing and
  evaluating several available tack methods.

  This paper investigates a dynamic weighting approach to choose the
  best method to perform a tack in order to minimise the number of
  failed tacking attempts whilst having minimal sensor knowledge of
  weather and boat state.
  After introducing the Black Python vessel, its sensors and the software
  structure in the Systems section, we focus on the software components that
  control the tack manoeuvre, suggesting several tacking implementations and a
  tack weighting process based on previous successful and failed tack manoeuvres.
  The methods introduced are demonstrated on experiment results obtained in
  coastal waters near Southampton.

\section{System presentation}
    \subsection{The boat} 
\begin{figure}[h]
    \centering
    \subfloat[The Black Python in Southampton water]{
    \includegraphics[width=0.41\textwidth]{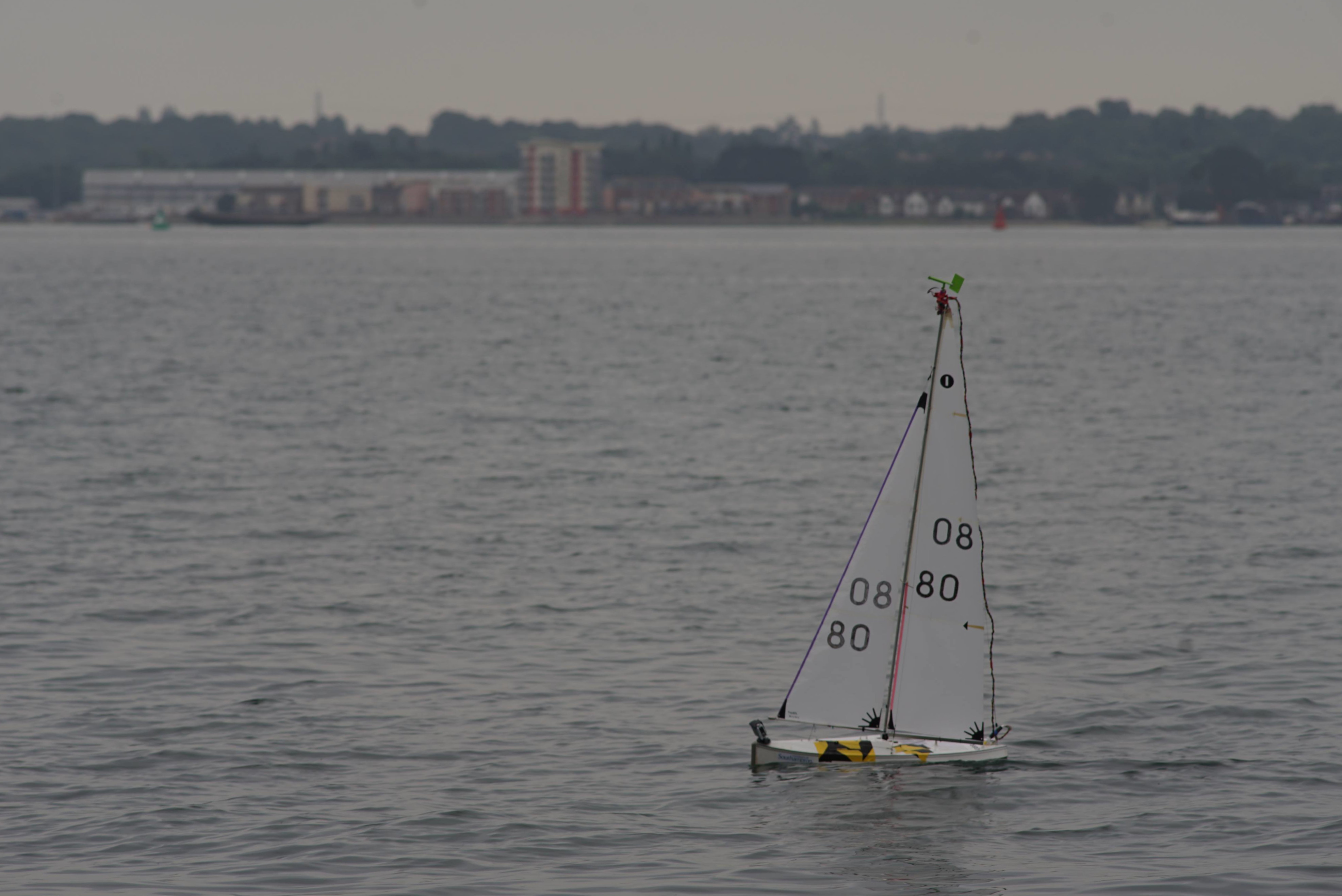}\label{fig:boat_lintel}}\hspace{0.02\textwidth}
    \subfloat[Profile view showing the center of effort of the
    smallest set of sails \cite{ilias2018}]{
    \includegraphics[width=0.25\textwidth]{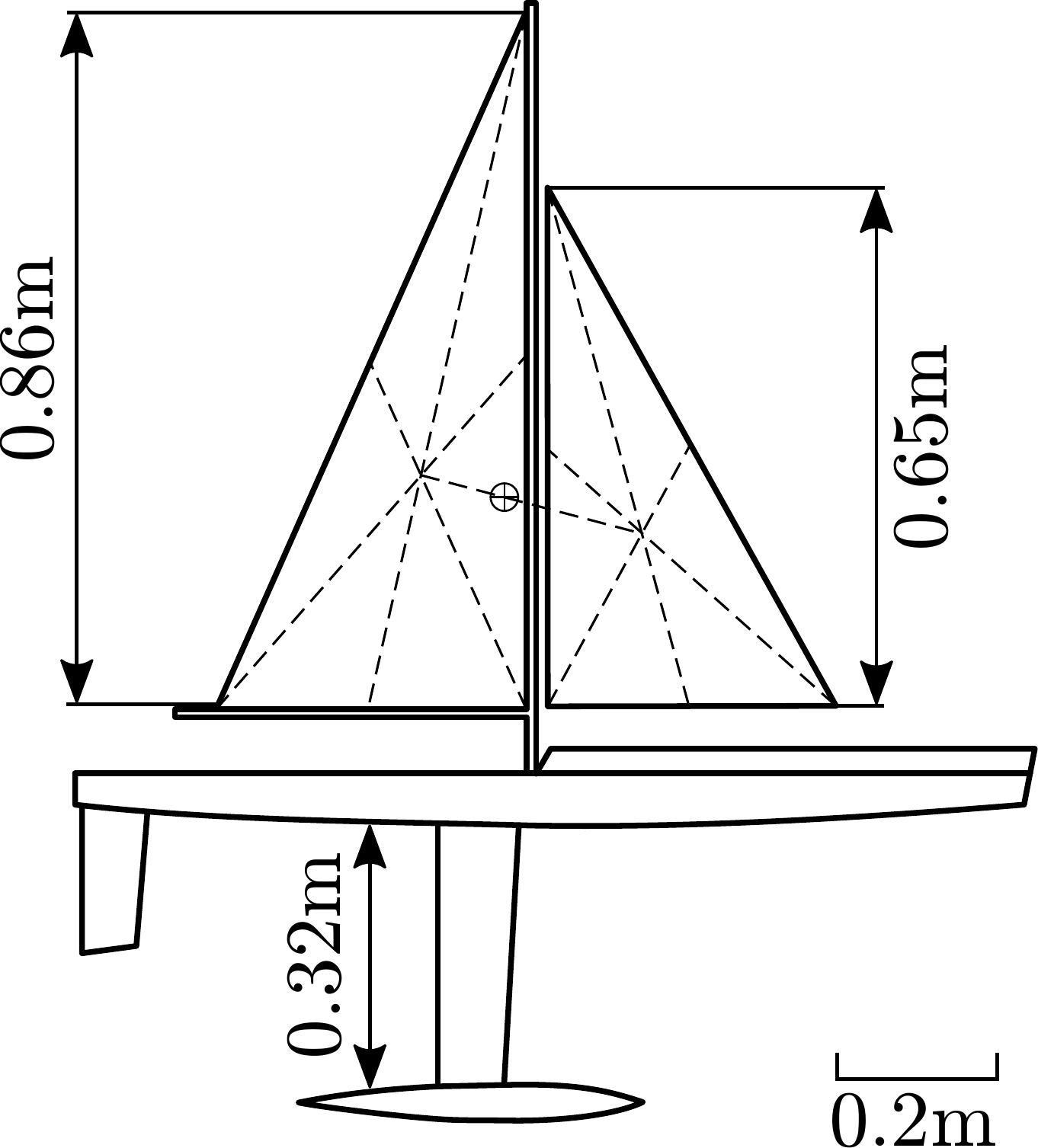}\label{fig:black_python_profile}}\hspace{0.02\textwidth}
    \subfloat[Custom made wind vane]{
    \includegraphics[width=0.26\textwidth]{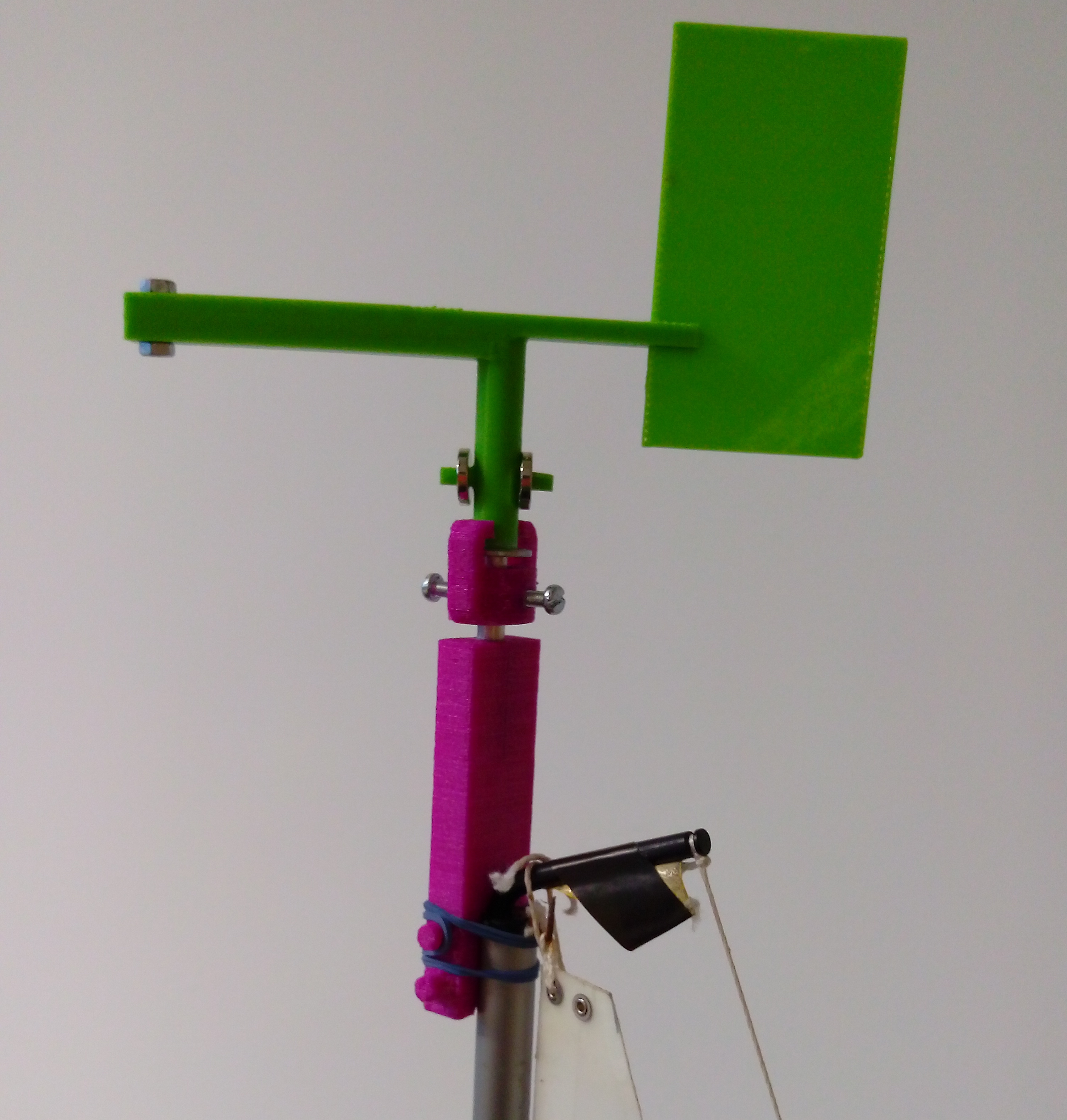}\label{fig:wind_vane_on_mast}}

    \caption{The Black Python}
    \label{fig:black_python}
  \end{figure}

  The Black Python, see Figure \ref{fig:boat_lintel}, is a
  one-meter-long Lintel mono-hull sailing robot yacht of class IOM
  (International One Metre) designed by David Creed. This boat is
  designed for racing performances and is used in remote controlled regatta.
  Three different sets of sails with a sail area of 6000, 4100 and
  \SI{2700}{\centi\meter\squared} can be used depending on the wind
  conditions. The hull's beam is \SI{165}{\milli\meter} and the hull
  displacement is \SI{4000}{\gram}.
  Minor changes have been made to fit
  wiring. The Black Python uses bulb keel that is \SI{420}{\milli\meter}
  deep, and has a spade rudder for steering.
  Profile view of the Black Python is shown in Figure
  \ref{fig:black_python_profile}.

    \subsection{Electronics}
  A Raspberry Pi 3 B (RPi) microcomputer is used as the main control
  board. It is powered by a 5V USB power bank. A foam stand and
  plastic box keep the RPi away from water that occasionally gets inside
  the boat.

  A uBLOX MAX M8Q GPS unit gives the boat position and velocity, it
  communicates with the RPi via I$^2$C (Inter-Integrated Circuit).  A
  conventional USB WiFi dongle is placed together with the GPS and a
  small IMU (Pololu AltIMU-10 v4) unit on the top of the mast.
  Inside the boat, an Xsens MTi 3 IMU (Inertial Measurement Unit) is
  used. It includes an accelerometer, gyroscope and magnetometer.
  The IMU is placed directly on top of the RPi.
  The yacht uses a custom made wind vane (Figure
  \ref{fig:wind_vane_on_mast}). Two magnets are attached to the rotating
  part and a Pololu AltIMU-10 v4 is placed on the stator. The
  magnetometer on the IMU detects the change in the magnetic
  field as the wind vane is rotated by the wind, determining the wind direction
  relative to the boat.

  A Futaba S3003 servomotor is used to drive the
  rudder and a HiTec 785 HB winch servomotor drives both mainsail and jib at
  once. A remote control receiver
  as well as a multiplexer are used to take control of the
  sailing robot in case of emergency or during the launch and recovery
  phases. The motors, multiplexer and RC receiver are powered by 4 AA batteries
  which are monitored with an Adafruit INA219 current sensor to ensure
  sufficient power for a remote controlled recovery.

  \subsection{Software}
    As mentioned in the previous section, the main computer of the Black Python 
    is a Raspberry Pi. It runs the GNU/Linux distribution Ubuntu 16.04.
    The software is written in Python and utilises ROS (Robot Operating 
    System\footnote{\url{https://ros.org}}, \cite{quigley2009ros}); a framework for
    writing robot software. ROS includes a collection of tools and
    libraries to simplify the task of developing complex behaviours. In
    the ROS ecosystem, the code is structured around scripts called nodes. 
    Each node can send messages under a certain topic name: this
    is called \textit{publishing}. Nodes can also listen for specific
    topics by \textit{subscribing} to them. The entire software
    developed by the Southampton Sailing Robot Team is made available
    under the MIT free software
    licence\footnote{\url{https://github.com/Maritime-Robotics-Student-Society/sailing-robot}}.

    \begin{figure}[h]
      \centering
      \vspace{0.8cm}
      \includegraphics[width=1.00\textwidth]{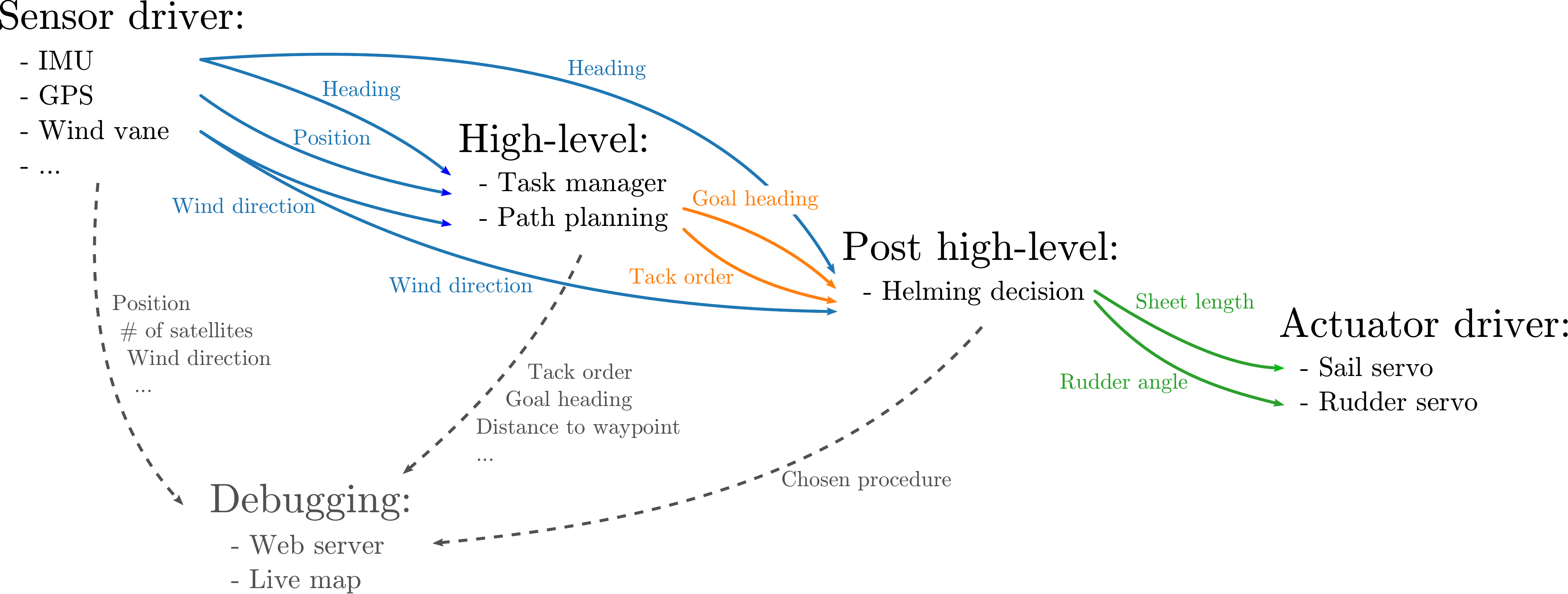}
      \caption{Structure of the code, arrows symbolise ROS messages
      passing}
      \label{fig:codestruct}
    \end{figure}

    \newpage
    The software of the Black Python is structured as follows:
    \begin{itemize}
      \item \textit{Drivers}: nodes talking directly to hardware components;
        this includes reading sensor data from the GPS, wind vane or obstacle avoidance camera
        as well as adjusting servo motor positions to set rudder angle and sail sheet. 
      \item \textit{High-level}: path planning nodes deciding when
        to switch tack and which heading to follow to complete a task. Available
        tasks are: reaching
        a waypoint, keeping a position, and avoiding an obstacle.
      \item \textit{Post high-level}: the \textit{helming node} converting the goal heading or tack
        order from high-level nodes into rudder angle and sail position.
        The \textit{helming node} is described in depth in section
        \ref{section:helming}.
      \item \textit{Debugging}: nodes for visualisation of messages; from maps
        with waypoint and boat positions to graphical displays of angle information like heading, goal
        heading, and wind direction
    \end{itemize}

    Figure \ref{fig:codestruct} illustrates the software structure, giving
    key nodes and the messages exchanging information between them.
    To achieve a particular assignment, the user configures the
    robot via parameter files. These include a list defining the tasks to run
    and task specific parameters like waypoint coordinates.

\section{Post high-level: dynamic tack control} \label{section:helming}

  Weather conditions are sometimes not suitable for tacking on such
  small boats (dues to waves for example). In this situations a reliable
  approach is to jibe instead. The switch between tacking and jibing to
  cross the wind was implemented in the past on the Black Python using a
  user defined parameter set before each test.  Whilst being a fail-safe
  method, we measured that jibing instead of tacking makes the vessel
  looses about 3m made good, when beating at 50 degree from the wind and
  with a speed of 0.75m/s. This leads to a loss of about 6 seconds per
  jibe, hence jibing should only happen when necessary.
  The post high-level was introduced to increase the choice of available
  manoeuvres alternatives to jibing and to automate the decision between
  all available tack and jibe manoeuvres.

  The post high-level layer currently consists of the \textit{helming
  node} alone, which operates between the low level drivers and the high
  level nodes. 
  When the boat is not trying to switch tack, the rudder angle is set
  using a heading PID control. To set the sail sheet length, a
  predefined look up table containing the apparent wind direction versus
  the sail sheet length is used. 
  When a tack order is given from the high-level nodes, the
  \textit{helming node} is responsible for implementing a successful tack by changing the
  sheet length and rudder angle demands, choosing from a set of
  available procedures based on past events and a dynamic weight system
  with an exploration component.

  In the context of the \textit{helming node}, a procedure is the name
  given to a series of instructions that commands the sail and the
  rudder to perform a change of tack (jibe or tack). Several procedures
  are implemented:
 
  \begin{itemize}
    \item Basic tack (\texttt{BasicTack}): the rudder is set to its maximum
      position either on the port side if the boat was sailing on a port
      tack, or on the starboard side otherwise.
    \item Basic jibe (\texttt{BasicJibe}): the rudder is set to its
      maximum position in reversed compared to a tack. To help with
      bearing away the sails are sheeted out. 
    \item Tack with sheet out (\texttt{TackSheetOut}): the rudder is set
      to its maximum position to perform a tack, and the sails are slightly
      sheeted out. On a conventional sailing boat the main
      sail tends to make the boat go more upwind, when the jib
      pushes the boat to go downwind. Sheeting out the jib can help with
      tacking, however on the Black Python both sails share the same
      control. This procedure hence tries to reduce the power in the jib by
      sheeting out a little both sails while conducting the tack. 
    \item Tack with speed build up (\texttt{TackIncreaseAngleToWind}):
      to speed up the boat and gain momentum to aid passing the tipping
      point of the tack the boat will bear away for 5 seconds at 80
      degrees from the wind. It will then perform an usual tack with setting the
      rudder to its maximum position. 
  \end{itemize}

  The basic functioning of the \textit{helming node} is as follows:
  Before each switch of tack, the procedure list is ordered by the time
  taken by each procedure in the past. The procedures in the list are
  tried in order until one succeeds to make the boat switch tack before
  a user defined timeout. After each procedure attempt, the time it
  took is recorded for future use to determine the order in the
  procedure list. A tack procedure is considered a success if the time
  the procedure took in order to have the boat on the opposite tack (at
  an angle between 50 and 120 degrees relative to the wind) is bellow
  the user defined \texttt{timeout}. 
  If a procedure fails it is placed further towards the end of the list
  by recording a value of 1.5 times the timeout. To ensure that all
  procedures are attempted, an exploration coefficient is considered as
  well.

  Three user defined variables are used:
  \begin{itemize}
    \item \texttt{timeout}: timeout in seconds after which a
      procedure is considered as failed
    \item \texttt{ProcedureList}: initial order of the procedure
      list
    \item \texttt{Exploration coefficient}: probability (0 to 1)
      of picking an untried procedure instead of the top list
      entry
  \end{itemize}

  The \texttt{ProcedureList} is a python list of dictionaries, each
  element of the list has three dictionary keys: \texttt{Procedure}
  which is a pointer to the procedure class, \texttt{TimeList}
  a list of the time taken by the last 10 attempts of this procedure,
  and finally \texttt{InitPos} the initial position of the procedure in
  the \texttt{ProcedureList} as defined by the user.

  Every time a tack is attempted, the \texttt{ProcedureList} is ordered
  based on weight. The procedure with the lowest weight will be tried
  first. The weights are computed as follows:

  If the procedure have been tried in the past (ie. \texttt{TimeList} is
  not empty), its weight is the mean of the elements of the
  \texttt{TimeList}, in other words the average time the procedure took
  in the past. On the other hand, if the procedure has never been
  tried before, the \texttt{TimeList} is empty, then the
  \texttt{Exploration coefficient} is used.  To know if the procedure
  will be picked by the exploration, a random number between 0 and 1 is
  generated. If it is above the \texttt{Exploration coefficient} no
  exploration is done.  A combination of the \texttt{timeout} and the
  \texttt{InitPos} is given as the weight. This places the procedure
  between already attempted procedures, before the failed ones but after
  the succeeded ones whilst keeping all unused
  procedures in order of the initial list.  Otherwise, if the randomly
  picked value is bellow the \texttt{Exploration coefficient} the
  procedure is placed at the top of the list by giving it a random
  weight between 0s and 0.1s (the random value ensures an arbitrary
  selection between methods picked by the exploration coefficient).
  The computation of the weight is summarised in Algorithm
  \ref{algo:getweight}.

  \begin{algorithm}
    \caption{Function to get the weight of each
    procedure}\label{algo:getweight}
    \begin{algorithmic}[1]
      \Function{getWeight}{\texttt{procedure}}
      \If {\texttt{procedure.TimeList} not empty}
      \State \Return mean(\texttt{procedure.TimeList})
      \Else
      \If {\texttt{random(0,1) $<$
      explore\_coef/number\_of\_untested\_procedures}}
      \State \Return \texttt{random(0, 0.1)}
      \Else
      \State \Return \texttt{timeout + 0.01*procedure.InitPos}
      \EndIf
      \EndIf
      \EndFunction
    \end{algorithmic}
  \end{algorithm}

  Once the \texttt{ProcedureList} is ordered, it will not be reordered
  until the next high-level command to change tack. The list entries are
  attempted in order until a procedure successfully completes before the
  timeout.  If all elements of the \texttt{ProcedureList} are tried
  without a success, the procedure selection will continue with the same
  list, beginning at the top entry.

  \subsection{Example of run}

\begin{figure}[H]
    \centering
    \includegraphics[width=0.4\textwidth]{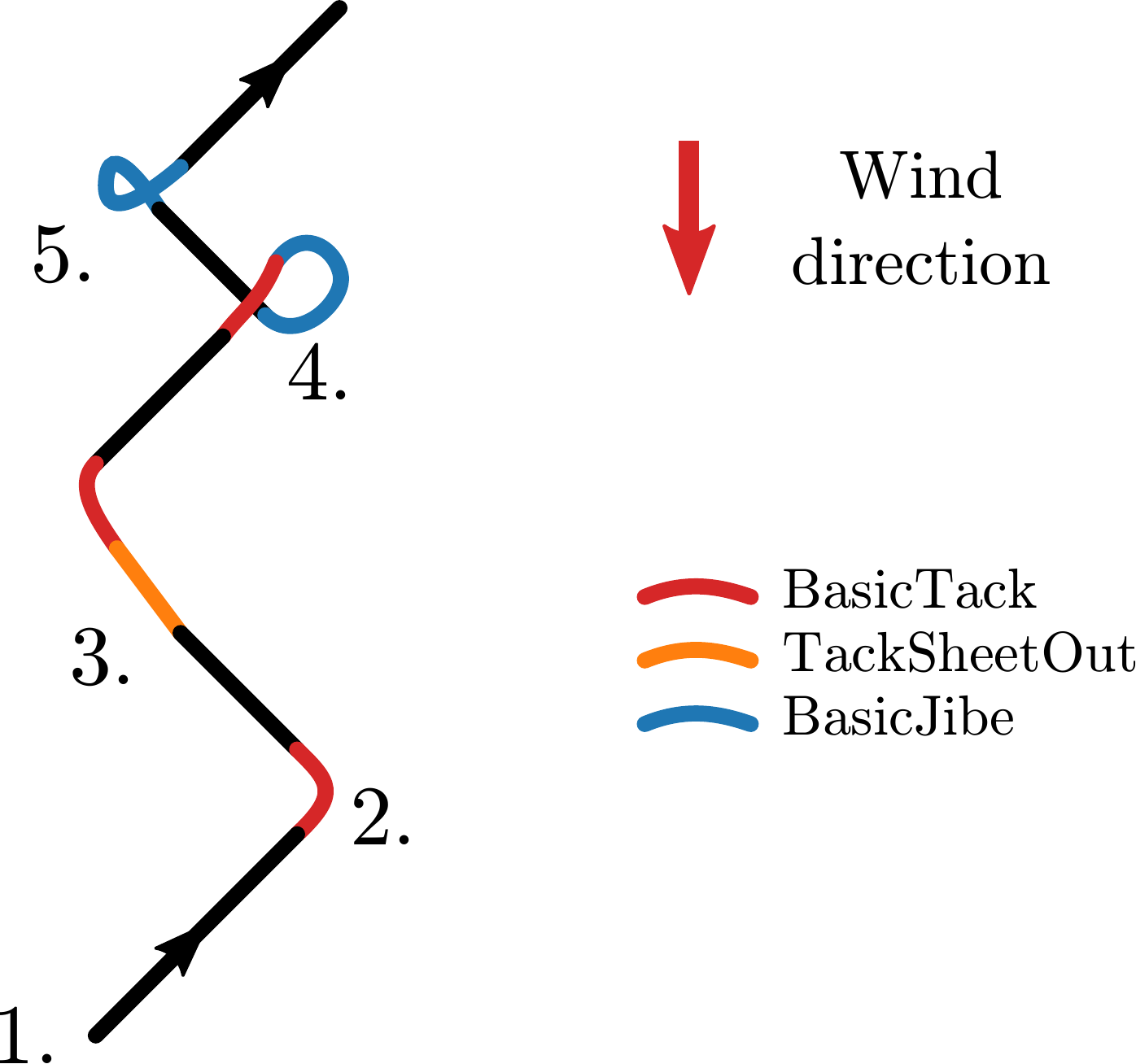}
    \caption{Fictional example of run with the \textit{helming node} procedure picking}
    \label{fig:helmingrun}
  \end{figure}

  In this section a step by step example of a fictional run is described
  and illustrated in figure \ref{fig:helmingrun}. The wind is coming
  from the North, and the boat is beating upwind.  The user defined
  parameters are as follows:
  \begin{itemize}
    \item \verb"Timeout": 15s
    \item \verb"ProcedureList": [\texttt{BasicTack}, \texttt{TackSheetOut}, \texttt{BasicJibe}]
    \item \verb"Exploration coefficient": 0.3
  \end{itemize}

  \begin{enumerate}
    \item The boat starts sailing on a port tack, close hauled.
    \item The \textit{helming node} receives a tack change command from the
      high-level nodes. The \texttt{ProcedureList} is sorted.
      None of the unused (all) procedures is moved to the top of the list
      through the \texttt{Exploration
      coefficient}, hence the \texttt{ProcedureList} is as
      defined by the user: [\texttt{BasicTack}, \texttt{TackSheetOut},
      \texttt{BasicJibe}].  The \texttt{BasicTack} is tried, it succeeds
      in 7 seconds. The boat continues on a starboard tack.

    \item The high-level commands the \textit{helming node} to change
      tack. The \texttt{ProcedureList} is ordered. This time the weight
      of the \texttt{BasicTack} is 7, making it the node with the fastest
      average time. The \texttt{Exploration
      coefficient} places the \texttt{TackSheetOut} procedure first in
      the \texttt{ProcedureList}:
      [\texttt{TackSheetOut}, \texttt{BasicTack}, \texttt{BasicJibe}].
      The \texttt{TackSheetOut} is tried, but does not succeed before
      the timeout of \SI{15}{\second}. A value of 22.5 (1.5 times the
      timeout) is stored in the \texttt{TimeList} for the
      \texttt{TackSheetOut}. The next procedure on the list, the
      \texttt{BasicTack}, is tried and succeeds in 8 seconds. The boat
      continues on a port tack.

    \item The high-level commands the \textit{helming node} to change
      tack. The \texttt{ProcedureList} is ordered, the \texttt{Exploration
      coefficient} causes no re-ordering. The
      \texttt{ProcedureList} is:
      [\texttt{BasicTack}, \texttt{BasicJibe}, \texttt{TackSheetOut}].
      The \texttt{BasicTack} is tried and fails, the next procedure
      (\texttt{BasicJibe}) is tried and succeed in 9 seconds. The boat
      continues its course.

    \item The high-level commands the \textit{helming node} to change
      tack. The \texttt{ProcedureList} is ordered, no re-ordering is caused by
      the \texttt{Exploration coefficient}. The \texttt{BasicTack} has a weight of 12.5 (mean
      of 7, 8 and 22.5), \texttt{TackSheetOut} has a weight of 22.5
      (failed once) and \texttt{BasicJibe} has a weight of 9. The order
      is hence [\texttt{BasicJibe}, \texttt{BasicTack},
      \texttt{TackSheetOut}]. The \texttt{BasicJibe} is tried and it is a
      success. 
  
  \end{enumerate}

  \subsection{Discussion on parameter selection}
  The user defined parameters were purposefully kept at a minimum and
  chosen to have an easily understandable meaning. The exploration
  coefficient may however demand some experience of the boat behaviour
  to be set properly. A low exploration coefficient should be used when
  the user is confident with his ordering of the procedure list or when
  he knows that very few tacks will be performed during the test. Hence
  finding the best possible manoeuvre is not as rewarding as finding a
  working manoeuvre. On the other hand when a larger number of tacks are
  to be expected, setting the exploration coefficient higher will help
  finding the most performant method. The timeout should be set at the
  minimum value that allows ones boat to perform a tack manoeuvre in the
  expected (or all) weather conditions. A too low timeout will lead to
  the helming node cycling through the procedures and keeping failing
  when a too high value will make the boat loose time when trying a procedure
  for the first time.

\section{Experiments}

\begin{figure}[h] 
      \centering
      \subfloat[]{
      \includegraphics[trim={0 0 0 1.2cm}, clip, height=0.27\textwidth]{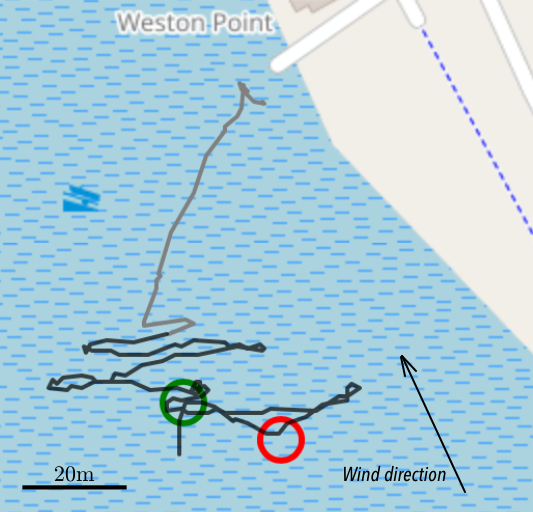}\label{fig:path_plot_a}}\hspace{1.1cm}
      \subfloat[]{
      \includegraphics[trim={0 0 0 1.2cm}, clip, height=0.27\textwidth]{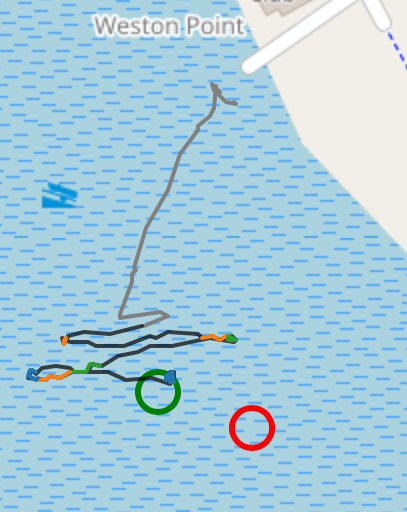}\label{fig:path_plot_b}}

      \subfloat[]{
      \includegraphics[trim={0 0 0 1.2cm}, clip,height=0.27\textwidth]{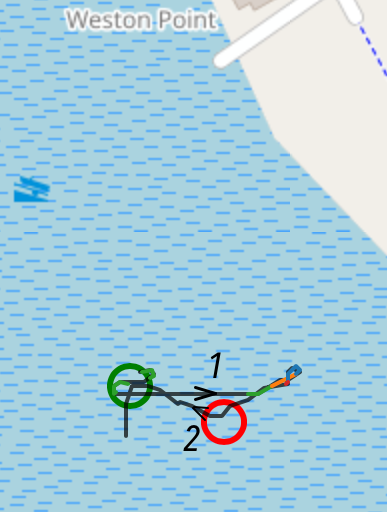}\label{fig:path_plot_c}}\hspace{1.1cm}
      \includegraphics[width=0.30\textwidth]{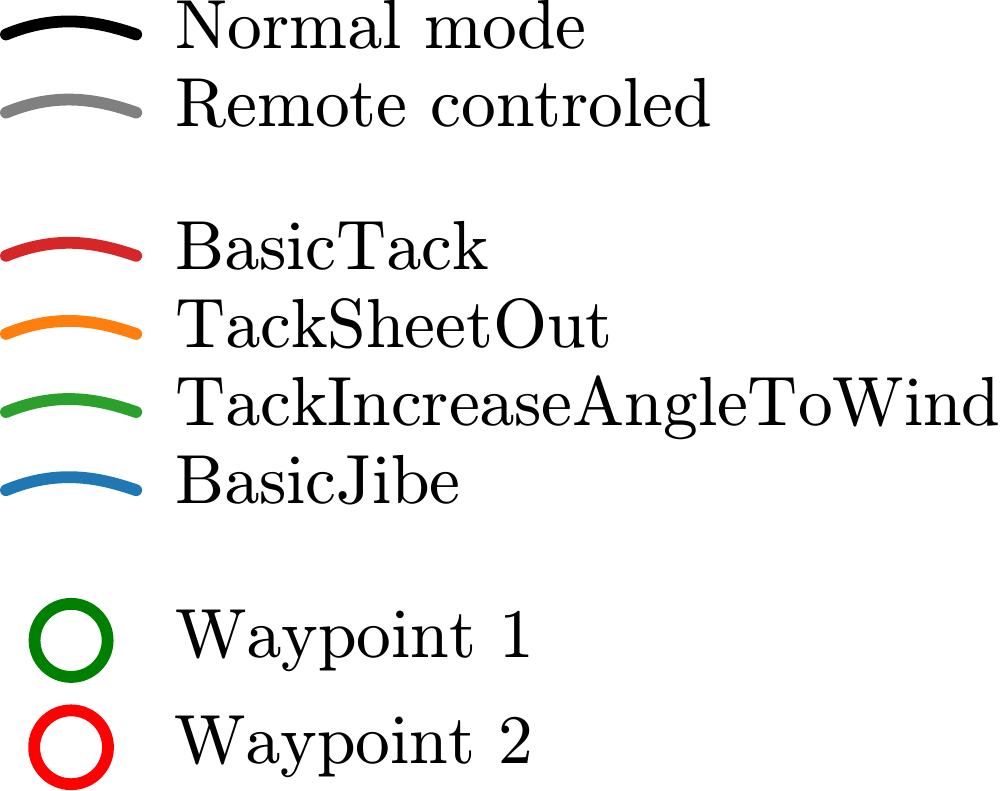}
      \caption{Experiment results in Southampton water}
      \label{fig:path_plot} 
    \end{figure}

The experiment was conducted on the sea near Southampton Sailing Club
shown in Figure \ref{fig:path_plot}. Wind direction gradually shifted from
south-east to south. The averaged wind direction according to the wind
direction data collected from the wind vane on the boat is shown in
Figure \ref{fig:path_plot_a}. Averaged wind speed was 4 knots with gusts
at 5 knots. Small waves with height of 15 - 20 cm have been observed during the
test. 

During the experiment, the boat was programmed to sail between two
waypoints separated by 20m shown as a green and red circle on Figure
\ref{fig:path_plot}.
The Black Python was released from a runway on north of the visible map.
To avoid being washed back by the waves the boat was first navigated
into the ocean by a remote control. Once the boat was further away from
the shore, the autonomous mode was activated. It tried to reach the
first waypoint using the \textit{helming node} described in previous section.
An acceptance radius of 1.5m was set for all waypoints, parameters used
in this experiment were:
\begin{itemize}
  \item \texttt{timeout}: 30s
  \item \texttt{ProcedureList}: [\texttt{BasicTack}, \texttt{TackSheetOut}, \texttt{TackIncreaseAngleToWind}, \texttt{BasicJibe}]
  \item \texttt{Exploration coefficient}: 0.3
\end{itemize}

Experiment results are shown in Figure \ref{fig:path_plot}. In the first
leg, three tack manoeuvres were made to reach the waypoint. 
As shown in Figure \ref{fig:path_plot_b}, the first
tack was done with a \texttt{TackSheetOut} procedure.
Whilst being the second element in the initial
\texttt{ProcedureList}, it was tried first because during the manual
controlled phase the \textit{helming node} was still running and a
\texttt{BasicTack} failed, moving this procedure at the end of the list.
This behaviour is not intended and will be fixed. The
\texttt{TackSheetOut} succeeded in 9s. For the second manoeuvre,
\texttt{TackSheetOut} was tried first and failed,
\texttt{TackIncreaseAngleToWind} was then conducted and succeeded in 19s.
For the next tack, \texttt{TackIncreaseAngleToWind} was tried first. It however failed, the next element in
the sorted list now being \texttt{TackSheetOut}. This procedure was
tried and also failed, finally a \texttt{BasicJibe} was conducted with
success in 19s.

After getting to the first waypoint, the boat sailed towards
the second waypoint as shown in Figure \ref{fig:path_plot_c}. Two
\texttt{TackIncreaseAngleToWind} were performed and succeeded. For the next
tack manoeuvre, all four procedures were tried:
first \texttt{TackIncreaseAngleToWind} did not manage to perform the
tack leading to the boat bearing away, then
\texttt{TackSheetOut} was tried without success. Later
\texttt{BasicJibe} was tried, the boat managed to switch tack, however,
the jibe did not finish
on time and switching to \texttt{BasicTack} was needed to finalise the
manoeuvre.
The last leg from the second waypoint to the first one is downwind, the
\textit{helming node} did not start any procedure and the boat
sailed in a straight line. 
After the boat reached the first waypoint again,
wind condition stopped us from doing any further repeating tests.

On this day, the low wind speed made it particularly difficult leading
to a lot of failed manoeuvres. Here the exploration never picked a
random untried procedure because all procedures were tried in the
first three tacks.
With such low wind conditions, increasing the timeout might help reducing
the number of failed manoeuvres because regardless of the method the boat is very slow
to switch heading.  This test demonstrated good functioning of the
\textit{helming node} and results show that in such weather jibing or
taking up speed by bearing away first helps switching tack.



\section{Concluding remarks and future work}
  Tacking on a small sailing boat is a delicate manoeuvre, the method presented
  here efficiently identifies a good strategy to perform a tack. 
  The key aspect of the \textit{helming node} framework and decision system is
  its simplicity; it does not rely on any additional hardware or even complex data
  processing but only on sensors that are already widely used on
  robotic sailing boats (heading and wind direction). It makes it easy to
  implement and to debug.
  From an user point of view, the simplicity of the
  implementation is also visible by the limited number of parameters and
  their physical meaning. Only 3 user defined parameters are needed
  (a timeout, a sorted list of procedure and an exploration coefficient) and all of them have an
  easy to understand meaning, no extensive knowledge of the boat
  behaviour or the weather conditions is needed.

  The \textit{helming node} system was successfully demonstrated, but it
  can still be improved on several aspects. If all the manoeuvres of the
  \texttt{ProcedureList} fail, this can mean that the selected timeout
  is too short. Instead of rerunning the list with the same parameter,
  increasing the timeout automatically would be judicious.  Additional
  tack procedures could further consider the sea state, for example by
  timing the tacks based on the position of the boat on the waves.
  Although some work to detect wave period has been done by the
  Southampton Sailing Robot Team, more tests are still needed to refine
  the method and integrate it as a procedure. 
  Lastly, for now only the time taken by a procedure is measured to
  assess its performances. To more accurately consider the distance
  lost during the jibe manoeuvre combining the time with the distance
  gained towards the next waypoint, for example, could be an improvement
  of the \textit{helming node} weighting process. Also the method is
  currently not suitable for long term tests where weather conditions
  might change over time. An improvement of the weighting that includes
  an aging parameter would be preferable in this case.
  
  After performing more tests with this new system, a
  better understanding of each procedure will be gained and 
  more precise and general conclusions concerning the best way to
  perform a tack in a specific weather condition can be drawn.


\newpage

\bibliographystyle{apalike} 
\bibliography{library.bib}

\begin{thebibliography}{}

\bibitem[Augenstein et~al., 2017]{augenstein2017using}
Augenstein, T., Singh, A., Miller, J., Pomerenk, A., Dean, A., and Ruina, A.
  (2017).
\newblock Using a controlled sail and tail to steer an autonomous sailboat.
\newblock In {\em Robotic Sailing 2016}, pages 91--103. Springer.

\bibitem[Cruz and Alves, 2014]{cruz2014navigation}
Cruz, N.~A. and Alves, J.~C. (2014).
\newblock Navigation performance of an autonomous sailing robot.
\newblock In {\em Oceans-St. John's, 2014}, pages 1--7. IEEE.

\bibitem[Gomes et~al., 2017]{gomes2017improving}
Gomes, L., Costa, A., Fernandes, D., Marques, H., and Anjos, F. (2017).
\newblock Improving instrumentation support and control strategies for
  autonomous sailboats in a regatta contest.
\newblock In {\em Robotic Sailing 2016}, pages 45--56. Springer.

\bibitem[Jouffroy, 2009a]{jouffroy2009control}
Jouffroy, J. (2009a).
\newblock A control strategy for steering an autonomous surface sailing vehicle
  in a tacking maneuver.
\newblock In {\em Systems, Man and Cybernetics, 2009. SMC 2009. IEEE
  International Conference on}, pages 2391--2396. IEEE.

\bibitem[Jouffroy, 2009b]{jouffroy2009steering}
Jouffroy, J. (2009b).
\newblock On steering a sailing ship in a wearing maneuver.
\newblock {\em IFAC Proceedings Volumes}, 42(18):26--31.

\bibitem[Langbein et~al., 2011]{langbein2011rule}
Langbein, J., Stelzer, R., and Fr{\"u}hwirth, T. (2011).
\newblock A rule-based approach to long-term routing for autonomous sailboats.
\newblock In {\em Robotic Sailing}, pages 195--204. Springer.

\bibitem[Masuyama and Fukasawa, 2011]{masuyama2011tacking}
Masuyama, Y. and Fukasawa, T. (2011).
\newblock Tacking simulation of sailing yachts with new model of aerodynamic
  force variation during tacking maneuver.
\newblock {\em SNAME Journal of Sailboat Technology}, 1.

\bibitem[Papadopoulos, 2018]{ilias2018}
Papadopoulos, I. (2018).
\newblock Techniques to improve the tacking performance of model scale robot
  yachts.
\newblock Technical report, University of Southampton.

\bibitem[Quigley et~al., 2009]{quigley2009ros}
Quigley, M., Conley, K., Gerkey, B., Faust, J., Foote, T., Leibs, J., Wheeler,
  R., and Ng, A.~Y. (2009).
\newblock {ROS}: an open-source robot operating system.
\newblock {\em ICRA workshop on open source software}, 3(3.2):5.

\bibitem[Roncin and Kobus, 2004]{roncin2004dynamic}
Roncin, K. and Kobus, J.-M. (2004).
\newblock Dynamic simulation of two sailing boats in match racing.
\newblock {\em Sports Engineering}, 7(3):139--152.

\bibitem[Spenkuch et~al., 2010]{spenkuch2010real}
Spenkuch, T., Turnock, S., Scarponi, M., and Shenoi, A. (2010).
\newblock Real time simulation of tacking yachts: how best to counter the
  advantage of an upwind yacht.
\newblock {\em Procedia Engineering}, 2(2):3305--3310.

\bibitem[Tranzatto et~al., 2015]{tranzatto2015debut}
Tranzatto, M., Liniger, A., Grammatico, S., and Landi, A. (2015).
\newblock The debut of aeolus, the autonomous model sailboat of {ETH} zurich.
\newblock In {\em OCEANS 2015-Genova}, pages 1--6. IEEE.

\bibitem[Tynan, 2018]{tynan2018attractor}
Tynan, D. (2018).
\newblock An attractor/repellor approach to autonomous sailboat navigation.
\newblock In {\em Robotic Sailing 2017}, pages 69--79. Springer.

\end{thebibliography}

\end{document}